\documentclass[lettersize,journal]{IEEEtran}
\usepackage{amsmath,amsfonts}
\usepackage{mathtools}
\usepackage{eqparbox}
\usepackage{bm}
\usepackage{algorithmic}
\usepackage{algorithm}
\usepackage{array}
\usepackage[caption=false,font=normalsize,labelfont=sf,textfont=sf]{subfig}
\usepackage{textcomp}
\usepackage{stfloats}
\usepackage{url}
\usepackage{verbatim}
\usepackage{graphicx}
\usepackage{cite}
\usepackage{float}
\usepackage{booktabs}
\usepackage{dsfont}
\usepackage{multirow}

\begin{document}

\title{Self-Supervised Depth Completion Guided by 3D Perception and
Geometry Consistency}
\author{Yu Cai, Tianyu Shen, Shi-Sheng Huang and Hua Huang,~\IEEEmembership{Senior Member, IEEE}
\thanks{This work was supported in part by the National Natural Science Foundation of China under Grant 62302047 and 62002020. (Yu Cai and Tianyu Shen contributed equally to this work.)

Y. Cai, T. Shen, S. Huang and H. Huang are with the School of Artificial Intelligence, Beijing Normal University, Beijing, 100875, China. Email: caiyu@mail.bnu.edu.cn, tianyu.shen@bnu.edu.cn, huangss@bnu.edu.cn, huahuang@bnu.edu.cn.

Corresponding author: Hua Huang.}}

\markboth{Journal of \LaTeX\ Class Files,~Vol.~14, No.~8, August~2021}%
{Shell \MakeLowercase{\textit{et al.}}: A Sample Article Using IEEEtran.cls for IEEE Journals}

\maketitle

\begin{abstract}
Depth completion, aiming to predict dense depth maps from sparse depth measurements, plays a crucial role in many computer vision related applications. Deep learning approaches have demonstrated overwhelming success in this task. However, high-precision depth completion without relying on the ground-truth data, which are usually costly, still remains challenging. The reason lies on the ignorance of 3D structural information in most previous unsupervised solutions, causing inaccurate spatial propagation and mixed-depth problems. To alleviate the above challenges, this paper explores the utilization of 3D perceptual features and multi-view geometry consistency to devise a high-precision self-supervised depth completion method. Firstly, a 3D perceptual spatial propagation algorithm is constructed with a point cloud representation and an attention weighting mechanism to capture more reasonable and favorable neighboring
features during the iterative depth propagation process. Secondly, the multi-view geometric constraints between adjacent views are explicitly incorporated to guide the optimization of the whole depth completion model in a self-supervised manner. Extensive experiments on benchmark datasets of NYU-Depth-v2 and VOID demonstrate that the proposed model achieves the state-of-the-art depth completion performance compared with other unsupervised methods, and competitive performance compared with previous supervised methods.
\end{abstract}

\begin{IEEEkeywords}
Depth completion, geometry consistency, 3D perceptual spatial propagation, self-supervised manner
\end{IEEEkeywords}

\section{Introduction}
\IEEEPARstart{D}{ue} to the hardware limitation and environmental interference, the depth information acquired by the existing depth sensors is frequently sparse, which cannot satisfy the requirements of the dense depth maps in many applications such as the 3D reconstruction\cite{refer1, refer64}, the virtual reality\cite{refer61, refer62, refer63}, the automatic driving\cite{refer4, refer5} and the robot map navigation\cite{refer6, refer7}. Therefore, it is of significance to study how to predict the dense and complete depth maps from the sparse depth measurements in the field of the computer vision.

In recent years, though the significant progresses have been made in the deep learning-based depth completion, achieving the high-precision depth completion results without relying on the ground-truth data still remains a challenge. On the one hand, the supervised methods\cite{refer18, refer19, refer22,refer13, refer14} heavily rely on the ground-truth data which can be high-cost and non-scalable in the practical applications. Additionally, the acquired data may be noisy, further complicating the process. On the other hand, the unsupervised methods\cite{refer27,refer28,refer29,refer30} eliminate the dependence on the ground-truth data, while the accuracy of unsupervised methods is still inferior to that of the supervised methods.

The quality of the depth completion results without the ground-truth data is insufficient due to the following two aspects. Firstly, the depth completion methods\cite{refer8} that solely consider the sparse depth information can only complete the depth maps with the simple structures and minimal gradient changes. They struggle to recover the edge details. Secondly, the depth completion methods\cite{refer9,refer10} guided by the RGB images have improved the quality of the completed depth maps, while they still fail to meet the accuracy requirements of many 3D visual tasks\cite{refer1,refer4,refer5,refer6,refer7,refer56}. At present, one of the optimal depth completion methods is the spatial propagation network (SPN)\cite{refer12} based on the semantically-aware affinity. However, most of the previous methods\cite{refer13,refer14} only exploit the 2D image features during the iterative propagation process, ignoring the 3D spatial structure and geometry information. 
This results in the unreasonable neighbor estimation and erroneous propagation, which easily causes the accumulation error and mixed-depth problem of the depth boundaries.

In order to alleviate the aforementioned problems, this study considers that integrating the 3D structural information into the deep model can provide favorable guidance for the
unsupervised depth completion. Compared to solely relying on the 2D image features, leveraging on the 3D structural information can result in more accurate and reasonable 3D spatial neighbor estimation, thereby aggregating more highly correlated neighbor features and suppressing the irrelevant features during the spatial propagation. Therefore, this study holds the potential to achieve a superior depth completion model and mitigate the mixed-depth problem at the depth boundaries by leveraging the 3D perceptual information and structural feature.

To achieve the above goals, we develop a high-precision self-supervised depth completion method based on the 3D perceptual spatial propagation mechanism and multi-view geometry consistency. Firstly, the paired RGB images and sparse depth measurement are inputted into an encoder-decoder network to acquire an initial depth map and affinity matrix. Then the initial depth map is used for constructing a point cloud representation, so as to reveal the 3D spatial structural information of the scenes. Finally, the 3D position information from the point cloud and acquired affinity matrix are integrated to guide the iterative spatial propagation for predicting the dense depth map. The paper evaluates the proposed method on the benchmark datasets of NYU-Depth-v2 and VOID. The main contributions of this paper are as follows:

1. The paper proposes a self-supervised depth completion method that integrates the 3D perceptual spatial propagation mechanism. By exploring the 3D information with a point cloud representation, the model can effectively perceive more relevant neighbors and accurately propagate more reliable depth information. Additionally, the multi-view geometry consistency module is utilized to guide the self-supervised model optimization.

2. The extensive experiments on the benchmark datasets of NYU-Depth-v2 and VOID demonstrate that the proposed method outperforms the most advanced unsupervised depth completion methods and achieves the competitive results compared to the previous supervised methods.

The rest of this paper is arranged as follows: In Section \uppercase\expandafter{\romannumeral2}, this part summarizes the related work in the field of the depth completion. Section \uppercase\expandafter{\romannumeral3} demonstrates the overall framework and detailed modules of our proposed method. In Section \uppercase\expandafter{\romannumeral4}, this part explains the experiments and comparative results of our model. Finally, Section \uppercase\expandafter{\romannumeral5} concludes this paper and presents the future directions in this field.

\section{Related Work}
\subsection{Supervised Depth Completion}

The supervised approaches based on deep learning have been extensively explored and effectively employed for the depth completion. Such learning-based supervised approaches aim to minimize a loss with respect to the ground-truth data.

Some early methods focus on learning morphology operators\cite{refer15} and compressed sensing\cite{refer16}. Dimitrievski et al.\cite{refer15} propose a morphological neural network, which integrates the morphological filtering operations to semantically enhance the completion quality of the depth information. Chodosh et al.\cite{refer16} employ an adaptive deep neural network (ADNN) based on the compressed sensing to extract the multi-level convolutional sparse features from the sparse depth, subsequently utilized for the depth prediction. The early methods are characterized by simplicity in implementation. However, due to the limitations of the morphology and compressed sensing information guidance, effectively handling the sparse depth remains a challenge. The early methods exhibit the suboptimal performance in the depth prediction.

Recently, the supervised depth completion methods mainly consider designing various network architectures and learning schemes to deal with the sparse depth effectively. The fusion architectures\cite{refer17,refer18,refer19} are proposed to process each modality separately. Apart from combining the multi-modal information, Chen et al.\cite{refer21} design a 2D-3D fusion network to to exploit the distinct dimensions of the feature information. In addition to fusing the feature information, some methods have been devised to harness the multi-scale information inherent in the sparse depth. The joint concatenation and convolution layers\cite{refer20} are performed for unsampling the sparse depth. Moreover, some methods not only focus on improving processing the sparse depth but also consider more effective ways to guide the optimization of the proposed depth completion models. Hu et al.\cite{refer11} fuse the independent images and depth networks to guide the spatial propagation network. Li et al.\cite{refer22} exploit a cascade hourglass network to handle diverse patterns during the depth completion effectively. In addition, the surface normals are utilized as the guidance to recover the depth regions with the large structural changes\cite{refer23,refer24,refer25}. Furthermore, the pixel confidence is capable of being computed and learned within a network\cite{refer26} to propagate the depth values with higher confidence levels for guiding the depth prediction optimization.

Although the recent supervised methods can realize high-quality depth completion results, they are extremely dependent on the ground-truth data for training, which typically are noisy and costly. Inspired by the method\cite{refer17},  the paper explores the utilization of the multi-view geometry consistency to realize the self-supervised training. 
\subsection{Unsupervised Depth Completion}
The mainstream unsupervised depth completion methods can be categorized into the algorithm-based traditional approaches and learning-based deep neural networks. The unsupervised methods assume that the additional data (stereo images or monocular videos) are available during training.

The algorithm-based traditional methods concentrate on the classical image processing algorithms that do not rely on any training data. Ku et al.\cite{refer8} propose a fast depth completion method by performing a closing operation on the small holes, followed by filling the large holes. However, Ku et al.\cite{refer8} predominantly leverage the sparse depth information, making it challenging to recover the depth information in the structurally complex regions. In order to optimize the depth completion guided by the additional information, Levin et al. design a depth completion method based on the colorization approach\cite{refer46} which poses the RGB information as the weighted guidance, along with the joint bilateral filtering method\cite{refer47} for denoising of the final depth map. Nevertheless, due to the absence of the model parameter optimization, such methods are typically limited to recovering the depth values in the regions with the simple structures and small gradients.

Recently, the learning-based unsupervised methods, which do not rely on the ground-truth data for optimizing the model parameters, have garnered increased research attention due to their ability to achieve the superior depth completion results. Both the stereo\cite{refer27} and monocular\cite{refer17,refer28,refer29,refer30} training paradigms focus on minimizing the photometric error between the current RGB image and its reconstructions from other views, which can be employed as a training supervised signal. However, due to the relatively weak supervised signal provided by only the photometric error, these models exhibit the subpar performance in recovering the depth values for the indoor scenes with the complex textures. To handle this problem, the additional synthetic datasets are leveraged for training. Yang et al.\cite{refer19} pretrain a separate network on an additional dataset, which learns the depth prior conditioned on the images. Rodriguez et al.\cite{refer48} utilize the supplementary synthetic datasets, which entail bridging the gap between the simulation and reality. Furthermore, an additional challenge in the depth completion is the sparsity, which renders the convolution largely ineffective as the activations in the early layers also tend to be close to zero. To obtain the dense depth prediction, some methods\cite{refer17,refer49,refer19} necessitate the adoption of the deep learning networks with the numerous layers and parameters. To address this issue, Wong et al.\cite{refer28,refer30} respectively infer a non-differentiable handcrafted grid approximation of the scene and a spatial pyramid pooling (SPP) method, which are susceptible to the detail error in the object regions with the sparse points or complex structures. Subsequently, Wong et al.\cite{refer31} also develop a depth completion method with the calibrated back-projection layers, which learns the trade-off between the density and detail to retain both the near and far structures. Besides, Liu et al.\cite{refer32} propose a monitored distillation method for the positive congruent depth completion, by exploiting the 2D feature information to predict the residual images. Such method balances the predicted distillation depth and unsupervised loss to guide the depth completion, achieving the optimal performance. Nonetheless, the 3D structural information are not explored effectively for such methods, which has the potential to misguide the depth completion.

The unsupervised methods eliminate the dependence on the ground-truth data compared to the supervised methods, while the depth completion accuracy of the existing unsupervised methods is still inferior to that of the supervised methods. To address the issue, the paper develops a self-supervised method that integrates the variant network of the SPN to effectively harness both the semantic and geometry information for guiding the pixel propagation, gaining the superior effectiveness. The proposed approach does not necessitate any supplementary data and is capable of propagating the valid depth values more effectively within the sparse depth maps.

\subsection{Spatial Propagation Networks}
Compared to the previous direct depth completion algorithms, leveraging the cross-modal information (e.g. RGB images and video) for the  propagation is a more efficient method to obtain the dense prediction from the depth sparse input. The current state-of-the-art methods is the SPN and variant approaches, where the RGB images and affinity matrix information are integrated to guide the depth completion.

Liu et al.\cite{refer12} propose the SPN to learn the semantically-aware affinity matrix, which is exploited in various high-level vision tasks. The SPN propagates in a row-wise and column-wise manner with a three-way connection, which only captures the local features in an inappropriate way. Cheng et al. \cite{refer13} present the convolutional spatial propagation network (CSPN) to resolve the ineffective issue of the SPN by performing the propagation process with a convolutional operation and predicting the affinity matrix of the local neighbors in four directions. Subsequently, Cheng et al.\cite{refer45} infer the CSPN++, building upon the foundation of the CSPN, to enable adaptive learning of the convolutional kernel sizes. The CSPN++ addresses the issue of the edge depth blurring that is present in the CSPN. However, the above methods only consider the fixed local neighbors, which potentially leads to the unreasonable neighbor estimation and erroneous propagation.

Recently, Park et al. \cite{refer14} develop a non-local spatial propagation network (NLSPN) to propagate the non-local correlated neighbors instead of the unrelated local neighbors. Despite the robustness of the NLSPN for estimating the non-local reasonable neighbors, it lacks the 3D geometric constraints and still is susceptible to estimating the erroneous neighbors, which result in the mixed-depth problem at the object boundaries.

In order to further alleviate the above issue and enhance the performance, the paper considers that the spatial propagation mechanism guided by the 3D perception can provide more favorable constraints for the unsupervised depth completion.

\begin{figure*}[htbp]
  \includegraphics[width=\textwidth]{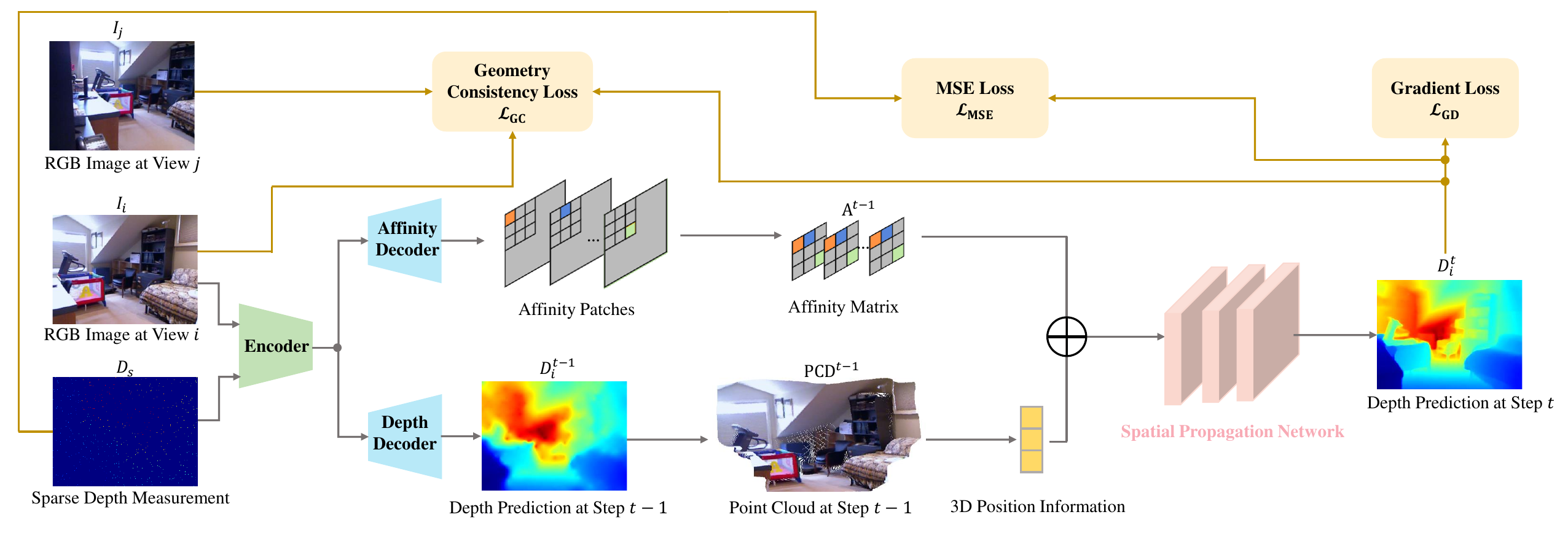}
  \caption{Overview of the proposed self-supervised depth completion method. Firstly the paired RGB images and sparse depth map are inputted into an encoder-decoder network to acquire the initial depth map and affinity matrix. Then the initial depth map is used for reconstructing the 3D point cloud. Finally, the 3D geometry information from the point cloud and acquired affinity matrix are integrated to guide the spatial propagation.}\label{fig1}
\end{figure*}

\section{Methods}
This section starts by introducing the overall architecture of the proposed approach that integrates the 3D perceptual spatial propagation mechanism and multi-view geometry consistency for guiding the model optimization. Then the details of each component and the comprehensive self-supervised loss function are demonstrated.

\subsection{Overall Architecture}\label{AA}
The proposed self-supervised framework based on the multi-view geometry consistency integrates a 3D perceptural spatial propagation network guided by an attention weighting mechanism, which can attain the higher-quality depth completion results. The overall architecture is illustrated in Fig. \ref{fig1}.

At first, an encoder-decoder network is utilized to jointly learn the initial depth map ${{D}_{i}^{0}}$ and affinity matrix $\mathbf{A}^{0}$. Then the 3D position information derived from the initial depth map ${{D}_{i}^{0}}$ concatenates the affinity matrix. During the iterative optimization phase, the concatenated information enters the spatial propagation network to guide the model optimization , ultimately yielding a dense depth prediction. For acquiring the current depth prediction ${{D}_{i}^{t}}$ at step $t$ ($t$ = 1, 2, 3...), the depth prediction ${{D}_{i}^{t-1}}$ is exploited to reconstruct the point cloud ${\rm PCD}^{t-1}$ ($t$ = 1, 2, 3...) with 3D position information. Then the affinity matrix $\mathbf{A}^{t-1}$ concatenates with the 3D position information. Finally, the spatial propagation with the attention weighting mechanism is performed to dynamically update ${{D}_{i}^{t-1}}$ to ${{D}_{i}^{t}}$, guided by the 3D position and affinity information. In the training stage, the self-supervised learning is achieved through an explicit incorporation of the geometric constraints between the adjacent views, while only a single-view RGB image and sparse depth measurement are required in the testing stage. 

\subsection{3D Perceptual Spatial Propagation}\label{AA}
Most of the previous spatial networks only consider the 2D image information as the guidance for the depth completion, causing the mixed-depth problem. Although the NLSPN method alleviates this issue by investigating the non-local neighbors in the 2D plane, disregarding the 3D structural information leads to the inaccurate neighbor estimation and erroneous spatial propagation. The Equation \ref{1} defines the neighbor estimation method of the previous spatial networks as follows:
\begin{equation}
N_{m,n}\!=\!\left\{\left(m\!+\!p, n\!+\!q\right)|\left(p,q\right)\in f\left({\mathbf{I}}\middle|\left(m,n\right)\right),p,q\in R\right\}\label{1}
\end{equation}
where $N_{m,n}$ denotes the neighbors of the pixel at position $\left(m,n\right)$. $f$ is the function to estimate the neighbors according to the information in the 2D plane, and $\mathbf{I}$ respresents the RGB images. For the local spatial propagation methods, the neighbor estimation function $f$ is fixed. For the original SPN\cite{refer12}, the information propagation occurs in a three-way connection, both in terms of rows and columns. In contrast, the CSPN\cite{refer13} achieves propagation by incorporating all local neighbors simultaneously in four directions. The CSPN++\cite{refer45} possesses the capacity to acquire the square kernels with diverse dimensions. Renowned as the most precise method previously, the NLSPN\cite{refer14} demonstrates the ability to comprehend the non-local neighbors within the 2D plane.

After extracting the 2D neighbors, the initial depth map can be generated through an encoder-decoder network or other networks, employing the intricate multi-modality fusion strategies. After several iterative steps, the final depth prediction is achieved, yielding more intricate and precise structural information. Each propagation step of the previous methods is formulated as follows:
\begin{equation}
d_{m,n}^{t+1}={\alpha_{m,n}^{m,n}d}_{m,n}^t+\sum_{(i,j)\epsilon N_{m,n}}{\alpha_{m,n}^{i,j}d}_{i,j}^t\label{2}
\end{equation}
where $t$ is the current iteration step, and $\left(m,n\right)$ denotes the coordinate of the center pixels in the 2D plane. $\left(i,j\right)$ indicates the coordinate of the neighbor pixels at position $\left(m,n\right)$, and $\alpha_{m,n}^{i,j}$ represents the affinity between the pixels at position $\left(m,n\right)$ and $\left(i,j\right)$. The affinity defines how much information should be passed between the adjacent pixels. Thus, each depth value is updated by the adjacent pixels according to the affinity. $d_{m,n}^{t+1}$ and $d_{m,n}^{t}$ respectively 
denote the depth value at position $\left(m,n\right)$ at the iteration step $t + 1$ and $t$. $d_{i,j}^{t}$ is the depth value at position $\left(i,j\right)$ at the iteration step $t$.

To explicitly integrate the geometric constraints into the spatial propagation, it is imperative to augment the neighbor estimation with the 3D structural information. This is achieved by projecting the depth prediction into the 3D space, in conjunction with the internal camera parameters denoted as $K$ in the Equation \ref{3}, following each propagation step.
\begin{equation}
\left[\begin{matrix}u\\v\\w\\1\\\end{matrix}\right]=d\left[\begin{matrix}{1/f}_x&0&-c_x/f_x&0\\0&{1/f}_y&-c_y/f_y&0\\0&0&1&0\\0&0&0&1\\\end{matrix}\right]\left[\begin{matrix}x\\y\\1\\1/d\\\end{matrix}\right]\label{3}
\end{equation}
where $\left(x,y\right)$ and $\left(u,v,w\right)$ respectively indicate the center coordinates of the patches in the 2D plane and projected 3D space; $d$ denotes the depth prediction during each propagation. $f_x$, $f_y$, $c_x$, and $c_y$ are the camera intrinsic parameters.
\begin{figure*}[htbp]
  \includegraphics[width=\textwidth]{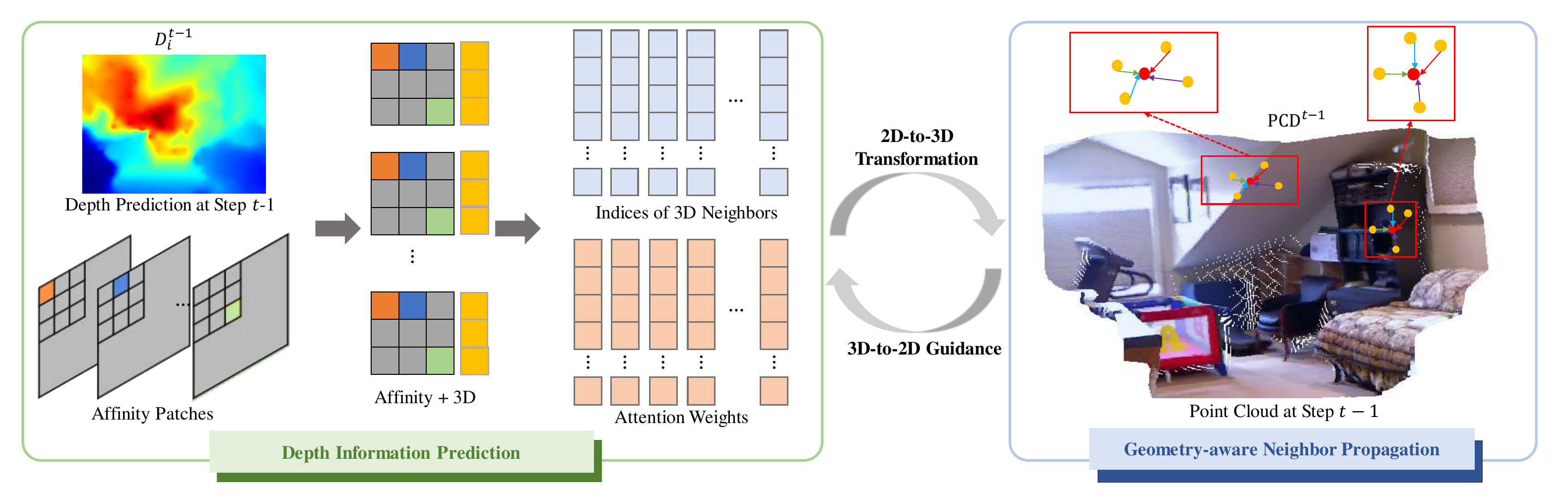}
  \caption{3D perceptual spatial propagation mechanism. Firstly, the 3D structural information and affinity matrix are exploited to achieve the 3D neighbors and attention weights of the neighbors. Then they are integrated to the geometry-aware neighbor propagation for the feature updating. The different colors of the edges indicate the different attention weights between the red center nodes and their yellow neighbors.}\label{fig2}
\end{figure*}

Upon obtaining the 3D representation, the k-nearest neighbor\cite{refer35} algorithm (KNN) is utilized as described in the Equation \ref{4} to identify the k-nearest neighbors within the 3D space surrounding the center patches. The selection of the k-nearest neighbors is determined through the calculation of the 3D Euclidean distances. Throughout each propagation iteration, each neighbor undergoes the reevaluation to mitigate the incorrect accumulation from the irrelevant neighbors.
\begin{equation}
\hat{N}\!=\!\left\{\left(u,v,w\right)|\left(u,v,w\right)\in F\left(\mathbf{D}\middle|\left(x,y\right)\right),x,y\in R\right\}\label{4}
\end{equation}
where $\hat{N}$ denotes the relative neighbors of the center pixel at position $\left(x,y\right)$, and $\left(u,v,w\right)$ represents the spatial coordinates of the neighbors in the 3D point cloud. The function $F$ is the KNN algorithm, which calculates the 3D Euclidean distances to estimate the neighbors. $\mathbf{D}$ indicates the depth prediction in the previous steps.

After extracting the 3D neighbors, we perform the spatial propagation by feature updating. 
In our paper, the attention weighting mechanism is integrated in the feature updating process. Note that the attention coefficients $\beta_{i,j}$ are calculated as follows:
\begin{equation}
\beta_{i,j}\!=\!\frac{{\rm exp}\ (\psi(x_i||x_j))}{\sum_{k\in \hat{N}^{t+1}(i)\cup{i}}{{\rm exp}\ (\psi(x_i||x_k))}}\label{5}
\end{equation}
where $\psi$ represents a parameterized multi-layer perception(MLP) neural network, and $| |$ denotes the concatenation operation. $x_i$ is the feature vector in the current position, and $x_j$ is the feature vectors of the neighbors. $\hat{N}^{t+1}(i)$ stands for the neighbors of the current pixel $i$ at the step ${t+1}$. The softmax function is applied for outputing the influence of each neighbor on the center pixel as the attention weights. Different attention weights assigned to the neighbors can effectively prevent the erroneous propagation of the unreasonable depth prediction during the iterative propagation.

Finally, the Equation \ref{6} and Fig. \ref{fig2} show the spatial propagation process:
\begin{equation}
x_i^{t+1}\!=\!\beta_{i,i}\varphi_i(x_i^t)+\sum_{j\epsilon \hat{N}^{t+1}(i)}{\beta_{i,j}\varphi_j(x_i^t||(x_j^t\ -\ x_i^t))}\label{6}
\end{equation}
where $x_i^t$ denotes the feature vector of the current pixel $i$ at the propagation step $t$, and $x_j^t$ stands for the feature vector of the neighbor pixel $j$ at the propagation step $t$. The KNN algorithm, guided by the 3D information, is utilized to estimate the neighbors. $\varphi$ is a neural network which is instantiated as a parameterized multi-layer perceptron (MLP). As a result, $x_i^t$ is updated into the feature vector $x_i^{t+1}$ at the step $t+1$.

\subsection{Multi-view Geometry Consistency}\label{AA}
\begin{figure*}[htbp]
  \includegraphics[width=\textwidth]{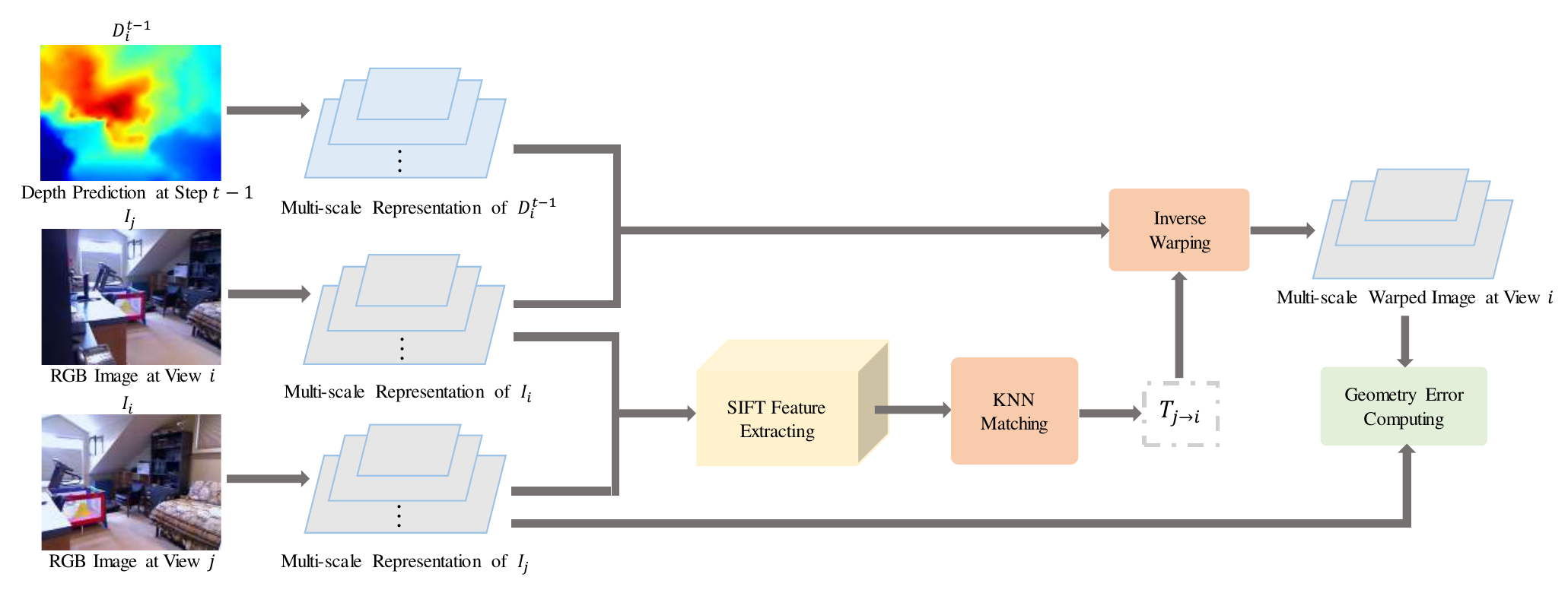}
  \caption{The pipeline of the multi-view geometry consistency module. $T_{j\rightarrow i}$ represents the transformation relationship between the current image and its adjacent images. For the RGB images at adjacent views, their multi-scale geometry error is calculated to realize the self-supervised training with the depth prediction.}\label{fig3}
\end{figure*}

To realize the self-supervised training, this study explores the utilization of the multi-view geometry consistency, as shown in the Fig. \ref{fig3}.

The calculation of the relative pose between the current frame and its adjacent frames is necessary, constituting an intermediary stage toward achieving the geometry consistency loss. The previous works have predominantly presumed the application of neural networks trained for the pose estimation. In contrast, within the proposed self-supervised framework, a model-based approach is implemented for the pose estimation, leveraging both the RGB image and sparse depth measurement.

The feature matching relation of ${I}_i$ and ${I}_j$ is established through the extraction of the scale invariant feature transform (SIFT) feature descriptors. The SIFT is capable of identifying the key points across the varying scales and determining their orientations. The salient characteristic of the SIFT-extracted key points lies in their resilience to the perturbing factors, including the variations in lighting and noise. Furthermore, the SIFT features manifest the notable stability as the local descriptors, remaining invariant to the transformations such as rotation, scaling, and alterations in brightness.

More specifically, the matched feature correspondences is leveraged to solve the Perspective-n-Point (PnP) problem\cite{refer34}, which enables to estimate the relative transformation $T_{j\rightarrow i}$ between the current frame and its adjacent frames. Then, the random sample consensus (RANSAC)\cite{refer33} method is combined with the solution of the PnP problem to further improve the robustness to the individual outliers during the feature matching process. This estimation offers the scale accuracy and an awareness of the potential failures, in contrast to the RGB-based estimation method\cite{refer44} which is up-to-scale.

Given the relative transformation $T_{j\rightarrow i}$ between the adjacent views, the current depth prediction ${{D}_{i}^{t}}$, and the camera's internal parameter matrix $K$, the adjacent image ${I}_j$ can be inversely warped to the current image. So for a certain pixel ${p}_j$ in the adjacent image ${I}_j$, ${p}_j$ has the corresponding projection ${p}_i$ in the current image ${I}_i$ in the Equation \ref{7}:
\begin{equation}
{p}_i\!=\!KT_{j\rightarrow i}{{D}_{i}^{t}}({I}_j)K^{-1}{p}_j\label{7}
\end{equation}

Consequently, the current corresponding RGB image can be acquired through applying the bilinear interpolation around the four neighbors of ${p}_i$.  In other words, the final warping process for all pixels ${p}_j$ is as follows:
\begin{equation}
{\rm warp}({I}_j\left({p}_j\right))\!=\!{\rm bilinear}({I}_i({p}_i))\label{8}
\end{equation}

The warped images exhibit similarity to the current RGB image under the static environmental condition, with the minimal occlusion attributable to the variations in the viewpoint. In the training stage, minimizing the error between the warped images and current image yields a reduction in the depth prediction error, contingent upon the proximity of the depth prediction to the ground-truth data.

If the projected point in the warped image differs by no more than 1 pixel from the true correspondence in the current image, it is considered to meet geometry consistency.

\subsection{Loss Function}\label{AA}
The proposed self-supervised framework operates independently of any ground-truth data. In the training stage, the self-supervised learning is realized by explicitly incorporating the geometric constraints between the adjacent views. Furthermore, the training procedure can be expedited by introducing the additional supervision in the form of a weighted auxiliary loss, which is applied to the output of each spatial propagation.

Firstly, the sparse depth measurement can serve as the form of a supervision signal. Specifically, the difference between the sparse depth measurement ${D}_{s}$ and depth prediction output ${D}_{i}^t$ on the set of the pixels with the valid depth is penalized. This loss results in higher accuracy and better stability. The Equation \ref{9} defines the MSE Loss function as follows:
\begin{equation}
\mathcal{L}_{\rm MSE}\!=\!\frac{1}{N_v}{\vert\vert{\mathds{1}_{D_{s} > 0}} \cdot {(D_{i}^t\!-\!D_{s})}\vert\vert}_{2}^{2}\label{9}
\end{equation}
where $N_v$ denotes the total number of the pixels with valid depth in the sparse input. ${\mathds{1}_{D_{s} > 0}}$ is an indicator function that equals to one if the depth value of ${D_{s}}$ is greater than 0, and zero otherwise. This loss function only calculates the difference of the valid depth values between the sparse depth input ${D_{s}}$ and the depth prediction output ${D_{i}^t}$.

Instead of punishing the depth difference like $L_{\rm MSE}$, the geometry consistency loss is computed exclusively based on the RGB pixel values, without the presence of the direct depth supervision.  The projected point exhibits the deviation from the true correspondence of no greater than 1 pixel. Therefore,  the multi-scale loss is computed by sampling half of each computed scale, ensuring that $\left|{p}_i^{\left(s\right)}-{p}_j^{\left(s\right)}\right|<1$ on at least one scale $s$. The geometry consistency loss function is defined as follows:
\begin{equation}
\mathcal{L}_{\rm GC}\!=\!\sum_{s\in S}\frac{1}{s}\left|{\mathds{1}}_{(D_{s}^{(s)}==0)} \cdot ({\rm warp}(I_j^{(s)})\!-\!I_i^{(s)})\right|\label{10}
\end{equation}
where $S$ represents the collection encompassing all scaling factors and $s$ denotes a scale in the set above. ${(.)}^{(s)}$ indicates the resized image (with average pooling) at the scale $s$. ${I_i^{(s)}}$ represents the current image at the scale s, and ${I_j^{(s)}}$ denotes the adjacent images at the scale s. ${{\rm warp}(I_j^{(s)})}$ is the warped RGB images from the adjacent views to the current view at the same scale. The weights of the losses associated with the lower resolutions are attenuated by the scale factor $s$. What's more, ${\mathds{1}}_{(D_{s}^{(s)}==0)}$ is an indicator function that equals to one if the depth value of $D_{s}^{(s)}$ is 0, and zero otherwise. This loss function only calculates the geometry consistency loss in the position where the depth value of $D_{s}^{(s)}$ is 0.

The previous two loss functions solely assess the aggregate of the individual errors computed independently on each pixel or depth, devoid of incorporating the neighbor constraints, which may lead to an unfavorable local optimum characterized by the pronounced discontinuity in the depth pixels. To mitigate this problem, the third gradient loss that calculates the second-order derivative of the depth prediction ${D}_{i}^t$ is proposed to promote the smoothness. The Equation \ref{11} defines the gradient loss as follows:\\
\begin{equation}
\mathcal{L}_{\rm GD}\!=\!\left|\nabla^2{D}_{i}^t\right|\label{11}
\end{equation}

In summary, the final loss function for the comprehensive self-supervised training framework ($\gamma_1$, $\gamma_2$ are the relative weight coefficients, and they are both assumed to be 0.1) is defined as follows:
\begin{equation}
\mathcal{L}_{\rm Self}\!=\!\mathcal{L}_{\rm MSE}+\gamma_1 \mathcal{L}_{\rm GC}+\gamma_2 \mathcal{L}_{\rm GD}\label{12}
\end{equation}

\begin{figure*}[htbp]
\centering
 \begin{minipage}{0.23\linewidth}
 	\vspace{3pt}
 	\centerline{\includegraphics[width=\textwidth]{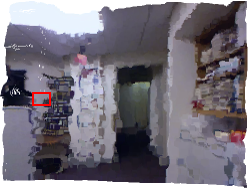}}
 	\vspace{1pt}
 	\centerline{\includegraphics[width=\textwidth]{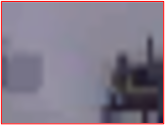}}
 	\vspace{3pt}
 	\centerline{SS-S2D}
 \end{minipage}
 \begin{minipage}{0.23\linewidth}
 	\vspace{3pt}
 	\centerline{\includegraphics[width=\textwidth]{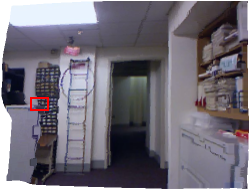}}
 	\vspace{1pt}
 	\centerline{\includegraphics[width=\textwidth]{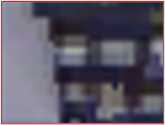}}
 	\vspace{3pt}
 	\centerline{Monitored Distillation}
 \end{minipage}
 \begin{minipage}{0.23\linewidth}
 	\vspace{3pt}
 	\centerline{\includegraphics[width=\textwidth]{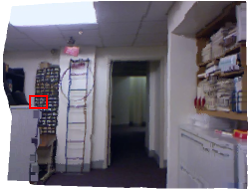}}
 	\vspace{1pt}
 	\centerline{\includegraphics[width=\textwidth]{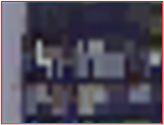}}
 	\vspace{3pt}
 	\centerline{Ours}
 \end{minipage}
 \begin{minipage}{0.23\linewidth}
 	\vspace{3pt}
 	\centerline{\includegraphics[width=\textwidth]{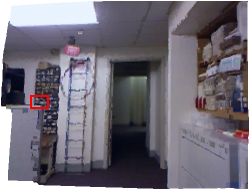}}
 	\vspace{1pt}
 	\centerline{\includegraphics[width=\textwidth]{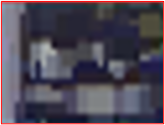}}
 	\vspace{3pt}
 	\centerline{Ground Truth}
 \end{minipage}
  \begin{minipage}{0.23\linewidth}
 	\vspace{3pt}
 	\centerline{\includegraphics[width=\textwidth]{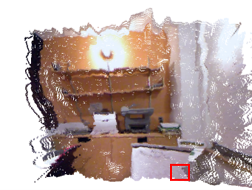}}
 	\vspace{1pt}
 	\centerline{\includegraphics[width=\textwidth]{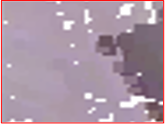}}
 	\vspace{3pt}
 	\centerline{SS-S2D}
 \end{minipage}
 \begin{minipage}{0.23\linewidth}
 	\vspace{3pt}
 	\centerline{\includegraphics[width=\textwidth]{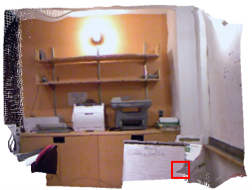}}
 	\vspace{1pt}
 	\centerline{\includegraphics[width=\textwidth]{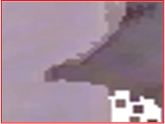}}
 	\vspace{3pt}
 	\centerline{Monitored Distillation}
 \end{minipage}
 \begin{minipage}{0.23\linewidth}
 	\vspace{3pt}
 	\centerline{\includegraphics[width=\textwidth]{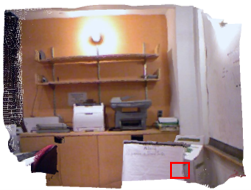}}
 	\vspace{1pt}
 	\centerline{\includegraphics[width=\textwidth]{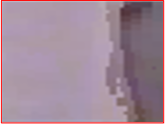}}
 	\vspace{3pt}
 	\centerline{Ours}
 \end{minipage}
 \begin{minipage}{0.23\linewidth}
 	\vspace{3pt}
 	\centerline{\includegraphics[width=\textwidth]{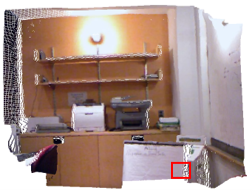}}
 	\vspace{1pt}
 	\centerline{\includegraphics[width=\textwidth]{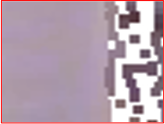}}
 	\vspace{3pt}
 	\centerline{Ground Truth}
 \end{minipage}
 \caption{Visual comparison of the 3D point cloud results on the NYU-Depth-v2 Dataset. The enlarged views of the red-boxed regions show that the proposed model recovers more accurate structures of the marginal details.}\label{fig4}
\end{figure*}

\section{Experiments}
The extensive experiments and ablation studies are conducted to verify the effectiveness of the proposed model. Firstly, the two datasets employed along with the quantitative evaluation metrics are introduced. Then in the intermediate content of this section, the implementation details are delineated and the qualitative and quantitative comparison of the proposed method with the state-of-the-art methods are conducted. Finally, the ablation studies are executed to assess the impact of each component.
\subsection{Datasets and Metrics}
\textbf{The NYU-Depth-v2 dataset}\cite{refer36} provides the RGB and depth images of 464 distinct indoor scenes, captured by using the Kinect sensor. The official dataset is utilized, with 249 scenes designated for training and remaining 215 scenes reserved for testing. And about 50K images from the training set are extracted for all experiments. Following the established testing protocol\cite{refer17}, the official test dataset comprising 654 images is employed to assess the ultimate performance of the whole model.

\textbf{The VOID dataset}\cite{refer30} encompasses the synchronized 640 × 480 RGB images and sparse depth maps, containing 56 scene sequence datasets. Among these sequences, 48 sequences, accounting for approximately 45,000 frames, are allocated for training, while the remaining 8 sequences, totally 800 frames, are reserved for testing. The corresponding sparse depth maps are categorized into three variants, namely 500 random samples, 1000 random samples, and 1500 random samples. These samples are derived from the collection of the features tracked by the visual-inertial odometry (VIO) system, specifically XIVO\cite{refer37}. The dense ground-truth data can be acquired through an active stereo imaging process.

 \textbf{Evaluation metrics}. The proposed method adopts some evaluation metrics from the previous works\cite{refer14, refer55}, including the mean absolute error (MAE [m]), the root mean square error (RMSE [m]), the inverse mean absolute error (iMAE [1/m]), the inverse root mean square error (iRMSE [1/m]), the absolute relative error (AbsREL) and the accuracy with threshold $t$ ($\delta_t$, for $t\in\left[1.25,\ {1.25}^2,{1.25}^3\ \right]$). 

\subsection{Implementation Details}
To implement the network, the PyTorch\cite{refer39} framework is used to conduct training on a single NVIDIA TESLA V100 GPU with 32 GB of memory. The encoder-decoder network is based on the residual network\cite{refer40}, generating the initial depth prediction and affinity patches. Subsequently, a three-layer multi-Layer perception (MLP) performs the 3D perceptual spatial propagation with the attention weighting mechanism.

For all experiments, the batch size is set to 12. The whole model adopts an ADAM optimizer\cite{refer41} with $\beta_1$ = 0.9, $\beta_2$ = 0.999 and is trained for 60 epochs. The initial learning rate is set to 0.0001, halved every 15 epochs.

For the pose-and-place (PnP) pose estimation, the adjacent images for the current frame are selected based on the nearest path. When the pose estimation encounters failure, the relative transformation $T_{j\rightarrow i}$ is assigned an identity matrix, and the current RGB image supersedes the adjacent images. Consequently, this circumstance yields a geometry consistency loss of 0, thereby exerting no influence on the training process.

For the sake of benchmarking against the state-of-the-art methods, both the NYU-Depth-v2 dataset and VOID dataset are utilized for both training and testing. For the NYU-Depth-v2 dataset, the original frames of size 640×480 are first downsampled to half, followed by the 304×228 center-cropping process, as seen in the previous works\cite{refer42, refer43}.
\subsection{Comparison with State-of-the-arts}

\begin{figure*}[htbp]
\centering
 \begin{minipage}{0.23\linewidth}
 	\vspace{3pt}
 	\centerline{\includegraphics[width=\textwidth]{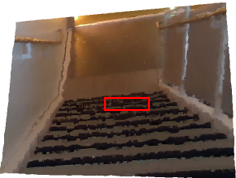}}
 	\vspace{1pt}
 	\centerline{\includegraphics[width=\textwidth]{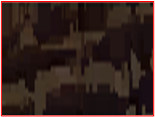}}
 	\vspace{3pt}
 	\centerline{SS-S2D}
 \end{minipage}
 \begin{minipage}{0.23\linewidth}
 	\vspace{3pt}
 	\centerline{\includegraphics[width=\textwidth]{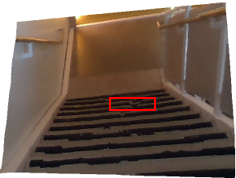}}
 	\vspace{1pt}
 	\centerline{\includegraphics[width=\textwidth]{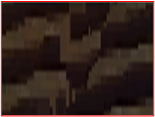}}
 	\vspace{3pt}
 	\centerline{Monitored Distillation}
 \end{minipage}
 \begin{minipage}{0.23\linewidth}
 	\vspace{3pt}
 	\centerline{\includegraphics[width=\textwidth]{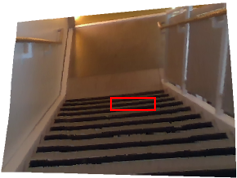}}
 	\vspace{1pt}
 	\centerline{\includegraphics[width=\textwidth]{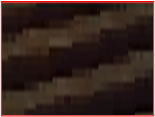}}
 	\vspace{3pt}
 	\centerline{Ours}
 \end{minipage}
 \begin{minipage}{0.23\linewidth}
 	\vspace{3pt}
 	\centerline{\includegraphics[width=\textwidth]{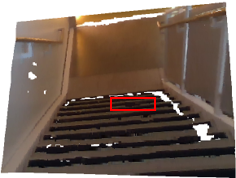}}
 	\vspace{1pt}
 	\centerline{\includegraphics[width=\textwidth]{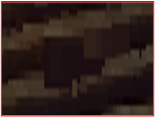}}
 	\vspace{3pt}
 	\centerline{Ground Truth}
 \end{minipage}
  \begin{minipage}{0.23\linewidth}
 	\vspace{3pt}
 	\centerline{\includegraphics[width=\textwidth]{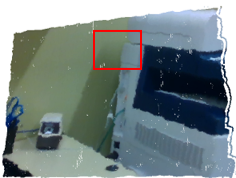}}
 	\vspace{1pt}
 	\centerline{\includegraphics[width=\textwidth]{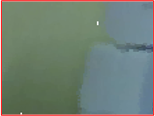}}
 	\vspace{3pt}
 	\centerline{SS-S2D}
 \end{minipage}
 \begin{minipage}{0.23\linewidth}
 	\vspace{3pt}
 	\centerline{\includegraphics[width=\textwidth]{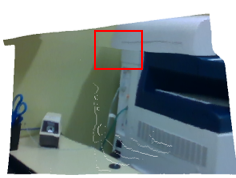}}
 	\vspace{1pt}
 	\centerline{\includegraphics[width=\textwidth]{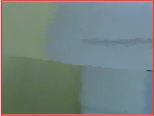}}
 	\vspace{3pt}
 	\centerline{Monitored Distillation}
 \end{minipage}
 \begin{minipage}{0.23\linewidth}
 	\vspace{3pt}
 	\centerline{\includegraphics[width=\textwidth]{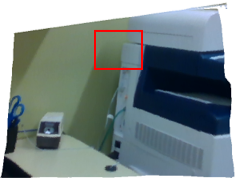}}
 	\vspace{1pt}
 	\centerline{\includegraphics[width=\textwidth]{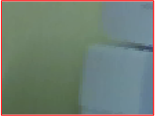}}
 	\vspace{3pt}
 	\centerline{Ours}
 \end{minipage}
 \begin{minipage}{0.23\linewidth}
 	\vspace{3pt}
 	\centerline{\includegraphics[width=\textwidth]{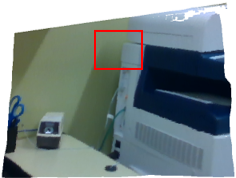}}
 	\vspace{1pt}
 	\centerline{\includegraphics[width=\textwidth]{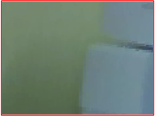}}
 	\vspace{3pt}
 	\centerline{Ground Truth}
 \end{minipage}
 \caption{Visual comparison of the 3D point cloud results on the VOID Dataset. The enlarged views of the red-boxed regions show that the edge structures recovered by the proposed method also achieve better results.}\label{fig5}
\end{figure*}

After acquiring the final depth prediction, the corresponding RGB image is projected into the 3D point cloud, which serves as the visual result. Then the 3D point cloud results of the whole model are evaluated against other state-of-the-arts methods on the two datasets mentioned above.

\begin{table*}[!ht]
  \centering
  \caption{The Comparison on the NYU-Depth-v2 dataset with other state-of-the-arts. The table compares against the previous unsupervised(U) and supervised(S) methods.}\label{table1}
    \begin{tabular}{cccccccccc}
    \toprule
    \multirow{2}{*}{Method} & \multirow{2}{*}{Type} & \multicolumn{5}{c}{Error Metric (lower is better)} & \multicolumn{3}{c}{Accuracy Metric (higher is better)} \\
\cmidrule{3-7}\cmidrule{8-10}    \multicolumn{1}{c}{} & \multicolumn{1}{c}{} & \multicolumn{1}{c}{MAE} & \multicolumn{1}{c}{RMSE} & \multicolumn{1}{c}{iMAE} & \multicolumn{1}{c}{iRMSE} & \multicolumn{1}{c}{AbsREL} & \multicolumn{1}{c}{$\delta<1.25$}   &  \multicolumn{1}{c}{$\delta<{1.25}^2$}   &  \multicolumn{1}{c}{$\delta<{1.25}^3$}\\
    \midrule
    SS-S2D\cite{refer17} & U     & \multicolumn{1}{c}{-} & 0.2710 & - & \multicolumn{1}{c}{-} & 0.0680 & \multicolumn{1}{c}{-} & \multicolumn{1}{c}{-} & \multicolumn{1}{c}{-} \\
    VOICED\cite{refer30} & U     & 0.1276 & 0.2284 & 0.0289 & 0.0547 & \multicolumn{1}{c}{-} & \multicolumn{1}{c}{-} & \multicolumn{1}{c}{-} & \multicolumn{1}{c}{-} \\
    ScaffNet\cite{refer28} & U     & 0.1175 & 0.1993 & 0.0249 & 0.0441 & \multicolumn{1}{c}{-} & \multicolumn{1}{c}{-} & \multicolumn{1}{c}{-} & \multicolumn{1}{c}{-} \\
    KBNet\cite{refer31} & U     & 0.1058 & 0.1978 & 0.0214 & 0.0427 & \multicolumn{1}{c}{-} & \multicolumn{1}{c}{-} & \multicolumn{1}{c}{-} & \multicolumn{1}{c}{-} \\
    Monitored Distillation & U     & \textbf{0.0779} & 0.1583 & 0.0146 & 0.0312 & 0.0252 & 0.987 & 0.997 & 0.999 \\
    \midrule
    S2D\cite{refer50} & S     & \multicolumn{1}{c}{-} & 0.2300  & \multicolumn{1}{c}{-} & \multicolumn{1}{c}{-} & 0.0440 & 0.971 & 0.994 & 0.998 \\
    S2D\cite{refer50}+Bilataral\cite{refer51} & S     & \multicolumn{1}{c}{-} & 0.4790 & \multicolumn{1}{c}{-} & \multicolumn{1}{c}{-} & 0.0840 & 0.924 & 0.976 & 0.989 \\
    S2D\cite{refer50}+ASAP\cite{refer52} & S     & \multicolumn{1}{c}{-} & 0.2320 & \multicolumn{1}{c}{-} & \multicolumn{1}{c}{-} & 0.0370 & 0.970  & 0.992 & 0.997 \\
    S2D\cite{refer50}+SPN\cite{refer12} & S     & \multicolumn{1}{c}{-} & 0.1720 & \multicolumn{1}{c}{-} & \multicolumn{1}{c}{-} & 0.0310 & 0.983 & 0.997 & 0.999 \\
    S2D\cite{refer50}+CSPN\cite{refer13} & S     & \multicolumn{1}{c}{-} & 0.1620 & \multicolumn{1}{c}{-} & \multicolumn{1}{c}{-} & 0.0280 & 0.986 & 0.997 & 0.999 \\
    S2D\cite{refer50}+UNet\cite{refer53} & S     & \multicolumn{1}{c}{-} & 0.1370 & \multicolumn{1}{c}{-} & \multicolumn{1}{c}{-} & \textbf{0.0200}  & \textbf{0.989} & \textbf{0.998} & \textbf{1.000} \\
    SS-S2D\cite{refer17} & S     & \multicolumn{1}{c}{-} & \textbf{0.1330} & \multicolumn{1}{c}{-} & \multicolumn{1}{c}{-} & 0.0270 & \multicolumn{1}{c}{-} & \multicolumn{1}{c}{-} & \multicolumn{1}{c}{-} \\
    \midrule
    \textbf{Ours}  & \textbf{U}     & 0.0802 & 0.1678 & \textbf{0.0122} & \textbf{0.0279} & 0.0282 & 0.982 & 0.997 & 0.999 \\
    \bottomrule
    \end{tabular}%
  \label{tab:addlabel}%
\end{table*}%

\begin{table}[htbp]
  \centering
  \caption{The Comparison on the VOID dataset with other state-of-the-arts. The table compares against the previous unsupervised(U) and supervised(S) methods}\label{table2}
  \scalebox{0.95}{
    \begin{tabular}{cccccc}
    \toprule
    \multirow{2}{*}{Method} & \multirow{2}{*}{Type} & \multicolumn{4}{c}{Error Metric (lower is better)} \\
\cmidrule{3-6}    \multicolumn{1}{c}{} & \multicolumn{1}{c}{} & \multicolumn{1}{c}{MAE} & \multicolumn{1}{c}{RMSE} & \multicolumn{1}{c}{iMAE} & \multicolumn{1}{c}{iRMSE} \\
    \midrule
    SS-S2D\cite{refer17} &  U     & 0.1789 & 0.2438 & 0.0801 & 0.1077 \\
    DDP\cite{refer19}   &  U     & 0.1519 & 0.2224 & 0.0746 & 0.1124 \\
    VOICED\cite{refer30} &  U     & 0.0851 & 0.1698 & 0.0489 & 0.104 \\
    VGG8\cite{refer30}  &  U     & 0.0985 & 0.1692 & 0.0572 & 0.1153 \\
    VGG11\cite{refer30} + SLAM\cite{refer54} &  U     & 0.0731 & 0.1464 & 0.0426 & 0.0932 \\
    ScaffNet\cite{refer28} &  U     & 0.0595 & 0.1191 & 0.0357 & 0.0684 \\
    KBNet\cite{refer31} &  U     & 0.0398 & 0.0959 & 0.0212 & 0.0497 \\
    Monitored Distillation\cite{refer32} &  U     & 0.0364 & 0.0878 & 0.0192 & 0.0438 \\
    \midrule
    ENet\cite{refer57}  &  S     & 0.0469 & 0.0944 & 0.0268 & 0.0526 \\
    MSG-CHN\cite{refer22} &  S     & 0.0436 & 0.1091 & 0.0234 & 0.0521 \\
    PENet\cite{refer11} &  S     & 0.0346 & 0.082 & 0.0189 & 0.0404 \\
    NLSPN\cite{refer14} &  S     & \textbf{0.0267} & 0.0791 & 0.0127 & 0.0339 \\
    \midrule
    \textbf{Ours}  &  \textbf{U}     & 0.0273 & \textbf{0.0709} & \textbf{0.0101} & \textbf{0.0199} \\
    \bottomrule
    \end{tabular}%
    }
  \label{tab:addlabel}%
\end{table}%

Firstly, the whole model is evaluated on the official test split data of the NYU-Depth-v2 dataset. The quantitative comparison results in the Table \ref{table1} demonstrate that the proposed approach exhibits the significant improvement when contrasted with the prior unsupervised approaches and delivers the competitive performance in the comparison to the earlier supervised methods, such as the SPN\cite{refer12} and CSPN\cite{refer13}. The method outperforms the state-of-the-art unsupervised method, Monitor Distillation\cite{refer32}, in terms of the iMAE and iRMSE metrics while exhibiting the comparable accuracy metrics. This superiority is attributed to the utilization of the 3D geometric constraints in the whole model, which provides the guidance for acquiring more accurate depth predictions. But the RMSE and MAE metrics of the monitored distillation slightly surpass those of the model, potentially due to the distillation process of the data removing outliers from the sparse depth measurements. The qualitative results presented in Fig. \ref{fig4} demonstrate the advantages of our approach in the depth completion of the object boundaries. The SS-S2D\cite{refer17} and monitored distillation\cite{refer32} don't consider the 3D geometric constraints and fail to capture the global relationships over the long distances, easily causing the mixed-depth problem at the object boundaries.

Secondly, the VOID dataset\cite{refer30} is utilized to further evaluate the ability of the method, as shown in the Table \ref{table2}. Notably, in terms of the RMSE, iMAE and iRMSE metrics, the proposed method surpasses the performance of the best unsupervised method Monitor Distillation\cite{refer32} and some previous supervised methods exhibiting favorable results, such as the PENet\cite{refer11} and NLSPN\cite{refer14}. In the scenarios characterized by distance variations within the uncomplicated environments, the 3D neighbors exhibits a higher degree of correlation compared to the 2D non-local neighbors chosen by the NLSPN. Consequently, the network of the method aggregates more highly correlated neighbors, which manifests greater precision in the depth completion. In the Fig. \ref{fig5}, the qualitative results of the proposed method are presented in the comparison to the state-of-the-art alternatives. In the context of the 3D representation, the model of this paper conspicuously achieves the superior visual effect in terms of the object shape accuracy compare to the SS-S2D\cite{refer17} and Monitor Distillation\cite{refer32}. This is because that suppressing the irrelevant neighbors can alleviate the issue of edge blurring caused by the significant gradient variations.

The improvement magnitude in various metrics on the VOID dataset is greater than that on the NYU dataset. The reason may be the difference in the complexity of scene structures between the two datasets. Given that the scenes on the VOID dataset primarily consist of the weakly textured areas with the limited 3D structural information, the model is inherently better positioned to acquire the comprehensive 3D geometry information for the model optimization guidance.

\subsection{Ablation Studies}
The ablation experiments are performed to demonstrate the effectiveness of each proposed component. The modified components of the whole model comprise: the 3D perception, the multi-view geometry consistency, the attention weighting mechanism, and the sparsity of depth samples.

\begin{figure*}[htbp]
\centering
 \begin{minipage}{0.23\linewidth}
 	\vspace{3pt}
 	\centerline{\includegraphics[width=\textwidth]{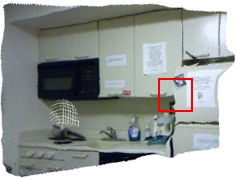}}
 	\vspace{1pt}
 	\centerline{\includegraphics[width=\textwidth]{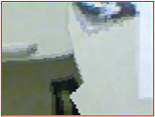}}
 	\vspace{3pt}
 	\centerline{3D Ablation}
 \end{minipage}
 \begin{minipage}{0.23\linewidth}
 	\vspace{3pt}
 	\centerline{\includegraphics[width=\textwidth]{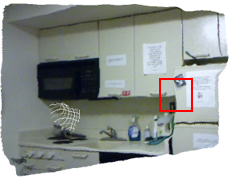}}
 	\vspace{1pt}
 	\centerline{\includegraphics[width=\textwidth]{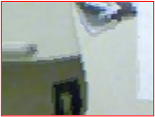}}
 	\vspace{3pt}
 	\centerline{Attention}
 \end{minipage}
 \begin{minipage}{0.23\linewidth}
 	\vspace{3pt}
 	\centerline{\includegraphics[width=\textwidth]{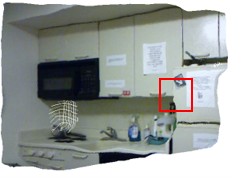}}
 	\vspace{3pt}
 	\centerline{\includegraphics[width=\textwidth]{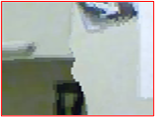}}
 	\vspace{3pt}
 	\centerline{Geometry Consistency}
 \end{minipage}
 \begin{minipage}{0.23\linewidth}
 	\vspace{3pt}
 	\centerline{\includegraphics[width=\textwidth]{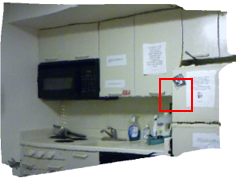}}
 	\vspace{1pt}
 	\centerline{\includegraphics[width=\textwidth]{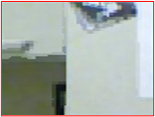}}
 	\vspace{3pt}
 	\centerline{Ours}
 \end{minipage}
  \begin{minipage}{0.23\linewidth}
 	\vspace{3pt}
 	\centerline{\includegraphics[width=\textwidth]{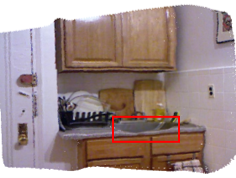}}
 	\vspace{1pt}
 	\centerline{\includegraphics[width=\textwidth]{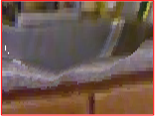}}
 	\vspace{3pt}
 	\centerline{3D Ablation}
 \end{minipage}
 \begin{minipage}{0.23\linewidth}
 	\vspace{3pt}
 	\centerline{\includegraphics[width=\textwidth]{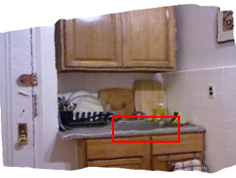}}
 	\vspace{1pt}
 	\centerline{\includegraphics[width=\textwidth]{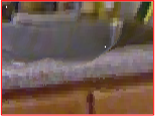}}
 	\vspace{3pt}
 	\centerline{Attention}
 \end{minipage}
 \begin{minipage}{0.23\linewidth}
 	\vspace{3pt}
 	\centerline{\includegraphics[width=\textwidth]{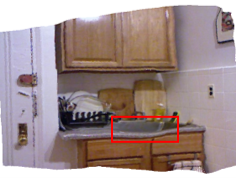}}
 	\vspace{3pt}
 	\centerline{\includegraphics[width=\textwidth]{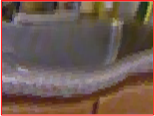}}
 	\vspace{3pt}
 	\centerline{Geometry Consistency}
 \end{minipage}
 \begin{minipage}{0.23\linewidth}
 	\vspace{3pt}
 	\centerline{\includegraphics[width=\textwidth]{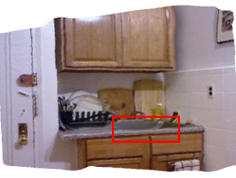}}
 	\vspace{1pt}
 	\centerline{\includegraphics[width=\textwidth]{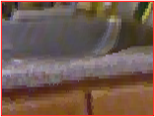}}
 	\vspace{3pt}
 	\centerline{Ours}
 \end{minipage}
 \caption{Visual comparison of the 3D point cloud results on the ablation experiments of each proposed component. With looking at the structural details in the enlarged views of the red-boxed regions, the results of the whole model are improved compared to those of the ablation experiments.}\label{fig6}
\end{figure*}

\textbf{3D perception.} The model of this paper explicitly incorporates the 3D structural features into the process of the spatial propagation. Upon disregarding the 3D geometric constraints, the neighbor estimation is confined solely to the feature space of the 2D image, lacking access to their respective spatial coordinates within the 3D space. As a result, the spatial propagation process may potentially disseminate the depth values of the irrelevant neighbors, resulting in distortion along the edges in the Fig. \ref{fig6}. The performance in the Table \ref{table3} also drops by a large margin which implies the importance of the 3D perception in the task of depth completion. The inefficacy is a consequence of notable discrepancies in predicting the depth values, particularly along the edges.

\begin{figure*}[htbp]
\centering
 \begin{minipage}{0.23\linewidth}
 	\vspace{3pt}
 	\centerline{\includegraphics[width=\textwidth]{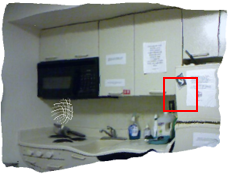}}
 	\vspace{1pt}
 	\centerline{\includegraphics[width=\textwidth]{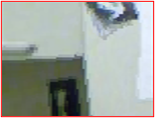}}
 	\vspace{3pt}
 	\centerline{200}
 \end{minipage}
 \begin{minipage}{0.23\linewidth}
 	\vspace{3pt}
 	\centerline{\includegraphics[width=\textwidth]{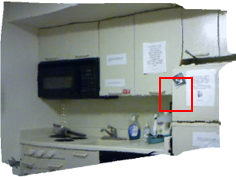}}
 	\vspace{1pt}
 	\centerline{\includegraphics[width=\textwidth]{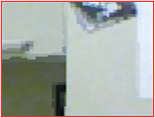}}
 	\vspace{3pt}
 	\centerline{500}
 \end{minipage}
 \begin{minipage}{0.23\linewidth}
 	\vspace{3pt}
 	\centerline{\includegraphics[width=\textwidth]{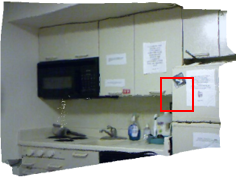}}
 	\vspace{1pt}
 	\centerline{\includegraphics[width=\textwidth]{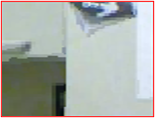}}
 	\vspace{3pt}
 	\centerline{800}
 \end{minipage}
 \begin{minipage}{0.23\linewidth}
 	\vspace{3pt}
 	\centerline{\includegraphics[width=\textwidth]{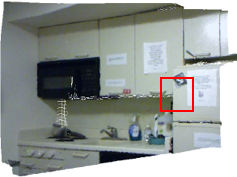}}
 	\vspace{1pt}
 	\centerline{\includegraphics[width=\textwidth]{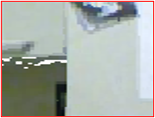}}
 	\vspace{3pt}
 	\centerline{Ground Truth}
 \end{minipage}
  \begin{minipage}{0.23\linewidth}
 	\vspace{3pt}
 	\centerline{\includegraphics[width=\textwidth]{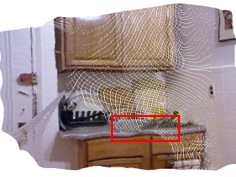}}
 	\vspace{1pt}
 	\centerline{\includegraphics[width=\textwidth]{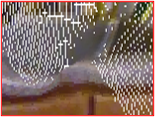}}
 	\vspace{3pt}
 	\centerline{200}
 \end{minipage}
 \begin{minipage}{0.23\linewidth}
 	\vspace{3pt}
 	\centerline{\includegraphics[width=\textwidth]{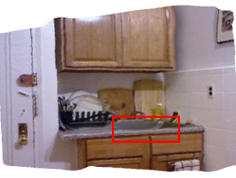}}
 	\vspace{1pt}
 	\centerline{\includegraphics[width=\textwidth]{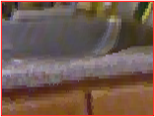}}
 	\vspace{3pt}
 	\centerline{500}
 \end{minipage}
 \begin{minipage}{0.23\linewidth}
 	\vspace{3pt}
 	\centerline{\includegraphics[width=\textwidth]{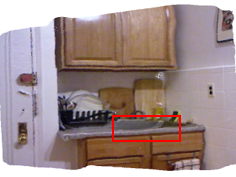}}
 	\vspace{1pt}
 	\centerline{\includegraphics[width=\textwidth]{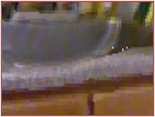}}
 	\vspace{3pt}
 	\centerline{800}
 \end{minipage}
 \begin{minipage}{0.23\linewidth}
 	\vspace{3pt}
 	\centerline{\includegraphics[width=\textwidth]{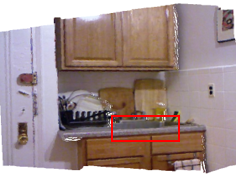}}
 	\vspace{1pt}
 	\centerline{\includegraphics[width=\textwidth]{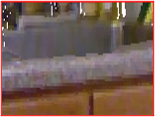}}
 	\vspace{3pt}
 	\centerline{Ground Truth}
 \end{minipage}
 \caption{Visual comparison of the 3D point cloud results (training on different input sparsity, testing on 500 random samples). While the performance varies as anticipated across different samples, the whole model consistently generates the results that are comparatively reasonable.}\label{fig7}
\end{figure*}

\textbf{Geometry consistency.} The reduction of the geometry error notably diminishes the error in depth prediction, specifically when the depth prediction closely aligns with the ground-truth data. To assess the essentiality of the geometry consistency, this experiment removes the geometry consistency module from the overall framework and solely penalizes the discrepancy in the valid depth. Consequently, the supervisory signal weakens in the presence of the limited valid depth within a sparse depth input, thereby deteriorating the performance of the RMSE metric, as evidenced in the Table \ref{table3} and Fig. \ref{fig6}. As shown in the Table \ref{table3} and \ref{table4}, the omission of the geometry consistency module reveals a scenario in which the evaluation metrics, specifically the MAE and AbsREL, exhibit the slightly improved performance. This may be attributed to the fact that the MAE and AbsREL reflect the true errors, whereas the RMSE magnifies the difference between the larger errors. Employing only the valid depth as the supervision signal effectively mitigates true errors. However, the geometry consistency loss exerts a higher penalty on larger errors, thereby optimizing the RMSE metric of the overall framework.

\textbf{Attention weighting mechanism.} As depicted in the Table \ref{table3} and Fig. \ref{fig6}, the performance of the overall framework suffers the significant decline in the absence of the attention weighting mechanism. This decline is attributed to the capability of the attention weighting mechanism to effectively hinder the erroneous spatial propagation. The attention weighting mechanism allocates the higher attention weights to more pertinent neighbors based on their affinity matrices, successfully mitigating the inherent over-smoothing issue in the edge details and resulting in more precise outcomes.

\begin{table*}[!ht]
  \centering
  \caption{The ablation studies about each component of Our method on the NYU-Depth-v2 dataset}\label{table3}
    \begin{tabular}{ccccccccccc}
    \toprule
    \multicolumn{1}{c}{\multirow{2}{*}{Attention}} & \multicolumn{1}{c}{\multirow{2}{*}{3D perception}} & \multicolumn{1}{c}{\multirow{2}{*}{Geometry consistency}} & \multicolumn{5}{c}{Error metric (lower is better)} & \multicolumn{3}{c}{Accuracy metric (higher is better)} \\
\cmidrule(r){4-8}\cmidrule(r){9-11}& \multicolumn{1}{c}{}& \multicolumn{1}{c}{}&\multicolumn{1}{c}{MAE} & \multicolumn{1}{c}{RMSE} & \multicolumn{1}{c}{iMAE} & \multicolumn{1}{c}{iRMSE} & \multicolumn{1}{c}{AbsREL} & \multicolumn{1}{c}{$\delta<1.25$}   &  \multicolumn{1}{c}{$\delta<{1.25}^2$}   &  \multicolumn{1}{c}{$\delta<{1.25}^3$}\\
    \midrule
    $\surd$     & $\surd$     & \multicolumn{1}{c}{} & \textbf{0.0799} & 0.1695 & 0.0121 & 0.0277 & \textbf{0.0281} & \textbf{0.982} & \textbf{0.997} & \textbf{0.999} \\
    $\surd$     & \multicolumn{1}{c}{} & $\surd$     & 0.0832 & 0.1721 & 0.0126 & 0.0290 & 0.0291 & 0.961 & 0.995 & 0.997 \\
    \multicolumn{1}{c}{} & $\surd$     & $\surd$     & 0.0836 & 0.1722 & 0.0125 & 0.0356 & 0.0291 & 0.981 & 0.997 & 0.999 \\
    \bm{$\surd$} & \bm{$\surd$} & \bm{$\surd$} & 0.0802 & \textbf{0.1678} & \textbf{0.0122} & \textbf{0.0273} & 0.0282 & \textbf{0.982} & \textbf{0.997} & \textbf{0.999} \\
    \bottomrule
    \end{tabular}%
  \label{tab:addlabel}%
\end{table*}%

\textbf{Robustness and generalization.} Adhering to the established protocol, the original model operates on the sparse depth input with 500 random samples. To assess the robustness of the proposed method, this experiment tests the method by using the sparse depth measurements, containing 200 samples and 800 samples. As shown in the Fig. \ref{fig7}, while the performance remains similar for models trained on the sparse depth measurements with 500 and 800 samples during testing on the sparse depth measurements with 500 samples, a substantial enhancement is evident when evaluating with 200 samples. As anticipated, the quantitative performance of the proposed method exhibits a decrease when dealing with 200 samples in the Table \ref{table5} and \ref{table6}, because of the reduction of the constraint information in self-supervised learning. However, it remained capable of generating the reasonably accurate results. Interestingly, when the input sparsity is transitioned to a higher level with 800 samples, the improvement in performance can be observed, albeit to a lesser extent compared to the effect of increasing the sample count from 200 (resulting in an RMSE metric of 0.2367) to 500 samples (resulting in an RMSE metric of 0.1678) in the Table \ref{table5}. The unevenness in the model optimization likely causes differential rates in the acquisition of the valid depth information. The results in the Table \ref{table5} and \ref{table6} emphasize the robustness and generalization capabilities of the whole model across the different input sparsity.

The results shown on the table \ref{table4} reveal that the removal of any one of the three components does not yield a significant improvement in the final evaluation metrics. This phenomenon may be attributed to the relatively simpler scenes and limited texture variations within the VOID dataset. Additionally, the selection of the reasonable neighbors in the 2D plane is feasible, and the impact of the structurally similar neighbors on the center patches appears nearly uniform.

\begin{table*}[!ht]
  \centering
  \caption{The ablation studies about each component of Our method on the VOID dataset}\label{table4}
    \begin{tabular}{ccccccccccc}
    \toprule
    \multicolumn{1}{c}{\multirow{2}{*}{Attention}} & \multicolumn{1}{c}{\multirow{2}{*}{3D perception}} & \multicolumn{1}{c}{\multirow{2}{*}{Geometry consistency}} & \multicolumn{5}{c}{Error metric (lower is better)} & \multicolumn{3}{c}{Accuracy metric (higher is better)} \\
\cmidrule(r){4-8}\cmidrule(r){9-11}& \multicolumn{1}{c}{}& \multicolumn{1}{c}{}&\multicolumn{1}{c}{MAE} & \multicolumn{1}{c}{RMSE} & \multicolumn{1}{c}{iMAE} & \multicolumn{1}{c}{iRMSE} & \multicolumn{1}{c}{AbsREL} & \multicolumn{1}{c}{$\delta<1.25$}   &  \multicolumn{1}{c}{$\delta<{1.25}^2$}   &  \multicolumn{1}{c}{$\delta<{1.25}^3$}\\
    \midrule
    $\surd$     & $\surd$     & \multicolumn{1}{c}{} & \textbf{0.0228} & 0.0714 & 0.0084 & 0.0225 & \textbf{0.0120} & \textbf{0.995} & \textbf{0.999} & \textbf{1.000} \\
    $\surd$     & \multicolumn{1}{c}{} & $\surd$     & 0.0240 & 0.0712 & 0.0090 & 0.0215 & 0.0128 & \textbf{0.995} & \textbf{0.999} & \textbf{1.000} \\
    \multicolumn{1}{c}{} & $\surd$     & $\surd$     & 0.0296 & 0.0713 & \textbf{0.0083} & 0.0217 & 0.0155 & \textbf{0.995} & \textbf{0.999} & \textbf{1.000} \\
    \bm{$\surd$} & \bm{$\surd$} & \bm{$\surd$} & 0.0273 & \textbf{0.0709} & 0.0101 & \textbf{0.0199} & 0.0147 & \textbf{0.995} & \textbf{0.999} & \textbf{1.000} \\
    \bottomrule
    \end{tabular}%
  \label{tab:addlabel}%
\end{table*}%

\begin{table*}[!ht]
  \centering
  \caption{The ablation studies about the number of the sparse depth input samples on the NYU-Depth-v2 dataset.}\label{table5}
    \begin{tabular}{ccccccccc}
    \toprule
    \multicolumn{1}{c}{\multirow{2}{*}{Sparsity}} & \multicolumn{5}{c}{Error Metric (lower is better)} & \multicolumn{3}{c}{Accuracy Metric (higher is better)}\\
\cmidrule(r){2-6}\cmidrule(r){7-9}&\multicolumn{1}{c}{MAE} & \multicolumn{1}{c}{RMSE} & \multicolumn{1}{c}{iMAE} & \multicolumn{1}{c}{iRMSE} & \multicolumn{1}{c}{AbsREL} & \multicolumn{1}{c}{$\delta<1.25$}   &  \multicolumn{1}{c}{$\delta<{1.25}^2$}   &  \multicolumn{1}{c}{$\delta<{1.25}^3$}\\
    \midrule
    200   & 0.1229 & 0.2367 & 0.0222 & 0.1962 & 0.0445 & 0.963 & 0.993 & 0.998 \\
    500   & 0.0802 & 0.1678 & 0.0122 & 0.0279 & 0.0282 & 0.982 & 0.997 & 0.999 \\
        800   & \textbf{0.0634} & \textbf{0.1429} & \textbf{0.0096} & \textbf{0.0234} & \textbf{0.0222} & \textbf{0.987} & \textbf{0.998} & \textbf{1.000} \\
    \bottomrule
    \end{tabular}%
  \label{tab:addlabel}%
\end{table*}%

\begin{table*}[!ht]
  \centering
  \caption{The ablation studies about the number of the sparse depth input samples on the VOID dataset.}\label{table6}
    \begin{tabular}{ccccccccc}
    \toprule
    \multicolumn{1}{c}{\multirow{2}{*}{Sparsity}} & \multicolumn{5}{c}{Error Metric (lower is better)} & \multicolumn{3}{c}{Accuracy Metric (higher is better)} \\
\cmidrule(r){2-6}\cmidrule(r){7-9}& \multicolumn{1}{c}{MAE} & \multicolumn{1}{c}{RMSE} & \multicolumn{1}{c}{iMAE} & \multicolumn{1}{c}{iRMSE} & \multicolumn{1}{c}{AbsREL} & \multicolumn{1}{c}{$\delta<1.25$}   &  \multicolumn{1}{c}{$\delta<{1.25}^2$}   &  \multicolumn{1}{c}{$\delta<{1.25}^3$}\\
    \midrule
    200   & 0.0418 & 0.1165 & 0.0239 & 0.0776 & 0.0235 & 0.987 & 0.997 & 0.998 \\
    500   & 0.0273 & 0.0709 & 0.0101 & 0.0199 & 0.0147 & 0.995 & 0.999 & 1.000 \\
    800   & \textbf{0.0164} & \textbf{0.0546} & \textbf{0.0062} & \textbf{0.0187} & \textbf{0.0089} & \textbf{0.997} & \textbf{0.999} & \textbf{1.000} \\
    \bottomrule
    \end{tabular}%
  \label{tab:addlabel}%
\end{table*}%

\section{CONCLUSION}
This paper proposes a self-supervised model that integrates the 3D perceptual spatial propagation mechanism for the sparse-to-dense depth completion. Different from the previous unsupervised methods, the proposed method incorporates the 3D geometry perception to more accurately estimate the relevant neighbors and the spatial propagation with an attention weighting mechanism to dynamically performed the feature aggregation and updating. The extensive experiments on the benchmark datasets of NYU-Depth-v2 and VOID demonstrate the effectiveness of our proposed method. The proposed method surpasses other unsupervised methods and attains the competitive results compared to some supervised methods. This underscores the robustness of the method of this paper, while the ablation studies reveal that the multi-view geometry consistency provides a relatively weak supervisory signal, resulting in only marginal enhancements of the RMSE metric. In the future, we intend to explore more effective strategies for enhancing unsupervised depth learning.

\end{document}